\documentclass[10pt,twocolumn]{article}

\usepackage[utf8]{inputenc}
\usepackage{lmodern}
\usepackage{amsmath,amssymb}
\usepackage{booktabs}
\usepackage{tabularx}
\usepackage{algorithm}
\usepackage{algorithmic}
\usepackage{enumitem}
\usepackage{microtype}
\usepackage{xspace}
\usepackage{xcolor}
\usepackage{pgfplots}
\usepgfplotslibrary{groupplots}
\usepackage{hyperref}
\usepackage[margin=0.75in]{geometry}

\hypersetup{
  colorlinks=true,
  linkcolor=blue,
  citecolor=blue,
  urlcolor=blue,
  pdfauthor={Huang Cheng, Scott Zhang, Aubert Li},
  pdftitle={FleetSieve: Decision-Critical Profiling for SLO-Aware LLM Fleet Configuration},
  pdfsubject={LLM serving systems},
  pdfkeywords={LLM serving, tensor parallelism, active profiling, SLO, fleet configuration}
}

\title{\Large\textbf{FleetSieve: Decision-Critical Profiling for\\SLO-Aware LLM Fleet Configuration}}
\author{Huang Cheng \qquad Scott Zhang \qquad Aubert Li\\
Meta, Menlo Park, California, USA\\
\texttt{huangcheng@meta.com}}
\date{}

\newcommand{\method}{FleetSieve\xspace}
\newcommand{\caplo}{\underline{c}}
\newcommand{\caphi}{\overline{c}}
\newcommand{\taillo}{\underline{\ell}}
\newcommand{\tailhi}{\overline{\ell}}

\pgfplotsset{compat=1.18}
\definecolor{fsblue}{HTML}{0072B2}
\definecolor{fsorange}{HTML}{E69F00}
\definecolor{fsgreen}{HTML}{009E73}
\definecolor{fsred}{HTML}{D55E00}
\definecolor{fspurple}{HTML}{CC79A7}
\definecolor{fsgray}{HTML}{777777}

\begin{document}
\maketitle

\begin{abstract}
Choosing tensor-parallel (TP) degrees and replica counts for an LLM serving fleet is difficult because performance is not monotonic in TP and the feasible choice can change with load. Exhaustive profiling resolves this uncertainty, but measures many configurations that do not affect the final resource allocation.

We present \method, which selects measurements according to their expected effect on a resource-coupled, SLO-aware fleet decision. \method models capacity and tail latency jointly, compares conservative and optimistic allocations, and stops when their remaining decision gap is below a specified tolerance. On a fixed H100 measurement grid for a 31B-parameter open-weight model, \method reaches the oracle aggregate decision using 22,200 GPU-seconds, 6.9\% less than uniform random profiling in the fixed comparison. Across 200 random reveal orders, its mean saving over random profiling is 5.4\% (95\% bootstrap CI: 3.5--7.2\%). The fixed-comparison saving is 21.5\% for Chat, while \method does not use the fewest GPU-seconds for Code. Joint capacity and tail modeling also avoids selecting a configuration whose 46.4-second completion p99 violates a 30-second SLO. In a 16-GPU allocation, an incorrect sparse-profile decision loses up to 1.93 requests/s and 12.4 percentage points of max-min fulfillment. Boundary repeats and BurstGPT measurements support the observed load-dependent tail-latency mechanism.
\end{abstract}

\section{Introduction}
\label{sec:introduction}

An LLM serving system rarely serves one homogeneous request stream. Interactive chat, code assistance, ranking, and background generation differ in arrival process, token lengths, and latency objectives. Under a finite GPU resource budget, an operator must choose a parallel configuration and a replica count for each class. These choices interact: assigning more GPUs to one replica may reduce its latency and increase its memory headroom, but it also leaves fewer GPUs for replicas serving other workloads.

Tensor parallelism illustrates this tension. Increasing TP shards model state over more GPUs and can improve per-request latency and KV-cache headroom, but it increases collective communication and reduces the number of replicas that fit in the same fleet. Consequently, neither \emph{always use the smallest TP} nor \emph{always use the largest TP} is generally correct. In our measurements, the SLO-feasible throughput per GPU peaks at the interior choice TP4 for both workloads. Under heavier chat load, however, TP4 violates the completion-time objective while TP8 remains feasible. Thus the relevant target is not a context-free TP ranking; it is the final resource-constrained fleet decision.

The conventional solution is to profile every candidate. This is reliable but scales with the Cartesian product of models, TP degrees, request shapes, loads, and serving policies. Generic active-learning and Bayesian-optimization methods reduce measurements, but they commonly target predictive uncertainty or the best isolated configuration. A fleet planner instead needs to know whether another measurement can change a coupled allocation.

\method treats profiling as downstream decision identification. It maintains uncertainty over capacity and tail latency, solves the fleet allocation under conservative and optimistic realizations, and measures configurations that are expected to reduce disagreement between those decisions. It stops only when critical feasibility is resolved and the remaining allocation gap is below a tolerance.

This paper makes three contributions:
\begin{enumerate}[leftmargin=*,nosep]
\item \textbf{Decision-critical acquisition.} Profiling is driven by a measurement's predicted effect on the downstream resource-coupled allocation rather than by grid coverage or marginal uncertainty alone.
\item \textbf{Joint capacity and tail modeling.} Capacity and the binding tail metric share the decision path, preventing a high-throughput but tail-infeasible configuration from being certified.
\item \textbf{A conditional decision certificate.} The profiler returns a three-state result---certified feasible, undecided, or certified infeasible---under an empirical uncertainty model. A replay that constructs priors from revealed measurements only produces the same stopping point and allocation.
\end{enumerate}

Our evaluation covers one 31B model and one H100 platform. \method reduces aggregate profiling GPU-seconds modestly, provides a larger reduction for Chat, and provides no reduction for Code. We therefore interpret the method as selectively useful when unresolved measurements can still change the allocation, rather than as a universal improvement.

\section{Background and Related Work}
\label{sec:related}

\paragraph{LLM serving.}
PagedAttention in vLLM~\cite{kwon2023vllm} and continuous batching in Orca~\cite{yu2022orca} improve memory utilization and scheduling. Sarathi-Serve~\cite{agrawal2024sarathi}, DistServe~\cite{zhong2024distserve}, Splitwise~\cite{patel2024splitwise}, and AlpaServe~\cite{li2023alpaserve} optimize batching, phase separation, or placement. SLOs-Serve~\cite{chen2025slosserve} allocates serving resources across multiple objectives, while Nitsum~\cite{srivatsa2026nitsum} adapts TP at runtime for tiered requests. These systems motivate a rich configuration space; \method addresses the complementary question of which measurements are needed before committing a fleet configuration.

\paragraph{Configuration search and decision-focused measurement.}
SCOOT~\cite{cheng2025scoot} applies constrained Bayesian optimization to LLM inference-engine tuning. Active learning more generally chooses informative samples~\cite{settles2012active}; targeted active learning explicitly optimizes information about a downstream Bayesian decision~\cite{filstroff2021targeted}. Fixed-confidence combinatorial exploration, such as CombGapE~\cite{nakamura2023combgape}, identifies an optimal combinatorial action with few samples. \method borrows this decision-focused perspective but instantiates it for correlated serving measurements, multiple SLO metrics, and an integer, resource-coupled fleet allocation. We do not claim that active learning, Bayesian optimization, elimination, or fixed-confidence stopping is independently new.

\paragraph{Public traces.}
Our primary replay uses the public Azure LLM inference traces accompanying DynamoLLM~\cite{stojkovic2025dynamollm}; the dataset is distributed under CC BY. BurstGPT~\cite{wang2024burstgpt} supplies an independent conversation trace for a mechanism-level robustness check.

\paragraph{Resource-constrained selection.}
Resource-constrained selection also appears in other infrastructure domains. Cheng et al.~\cite{cheng2014kcst}, for example, formulate continuous roadside coverage as a knapsack-constrained Steiner-tree problem. \method addresses a different system and uses a different algorithm, but shares the broader objective of concentrating a finite resource budget on decisions that determine system-level utility.

\section{Problem Formulation}
\label{sec:problem}

Let workload class $i\in\mathcal{W}$ have offered demand $\lambda_i$, an SLO $L_i$, and priority $p_i$. Priorities define a set of critical classes $\mathcal{C}$ and minimum fulfillment floors $\delta_i$ for $i\in\mathcal{C}$. A serving configuration $q\in\mathcal{Q}_i$ specifies a TP degree and an admission/load setting. It consumes $g_q$ GPUs per replica. Its unknown performance vector is
\[
  \theta_{iq} = (c_{iq}, \ell_{iq}, s_{iq}),
\]
where $c_{iq}$ is sustainable throughput, $\ell_{iq}$ is the relevant tail-latency metric, and $s_{iq}$ is success probability. A configuration is feasible only when its tail and success metrics satisfy the class policy.

The allocator chooses an integer replica count $n_{iq}$ and served load $x_i$. In our implementation each class uses one homogeneous configuration, expressed with a binary selector $y_{iq}$:
\begin{align}
  &\sum_q y_{iq}=1,\qquad n_{iq}\le M y_{iq},\\
  &x_i \le \sum_q n_{iq}c_{iq},\\
  &\sum_{i,q}n_{iq}g_q\le G,\\
  &0\le x_i\le\lambda_i,\quad n_{iq}\in\mathbb{Z}_{\ge0},\quad y_{iq}\in\{0,1\},\\
  &y_{iq}=1 \Rightarrow \ell_{iq}\le L_i,\; s_{iq}\ge s_i^{\min}.
\end{align}
The satisfaction ratio is $r_i=x_i/\lambda_i$. The objective is lexicographic: (1) enforce $r_i\ge\delta_i$ for critical classes whenever feasible, (2) maximize $\min_i r_i$, (3) maximize total served goodput, and (4) minimize fragmentation. Profiling uses a separate compute budget measured in GPU-seconds.

If all $\theta_{iq}$ were known, exhaustive profiling would produce the oracle allocation. The profiling problem is to approach that decision while revealing only a subset of the table.

\section{FleetSieve}
\label{sec:method}

\subsection{Joint uncertainty state}

After round $t$, \method maintains capacity intervals
$[\caplo_{iq,t},\caphi_{iq,t}]$ and tail intervals
$[\taillo_{iq,t},\tailhi_{iq,t}]$. The intervals combine an interior-knee structural prior with measured residuals. They are empirical rather than distribution-free; Section~\ref{sec:certificate} describes the audit that bounds this claim.

The conservative allocation $A_t^-$ uses lower capacity and upper tail bounds. The optimistic allocation $A_t^+$ uses upper capacity and lower tail bounds. Both use the same GPU constraint, SLO policy, and lexicographic objective.

\subsection{Decision certificate}

The state is \textsc{Certified-Infeasible} if even $A_t^+$ cannot meet the critical floor. It remains \textsc{Undecided} if critical feasibility is unresolved or the optimistic and conservative objectives differ by more than $\varepsilon$. It is \textsc{Certified-Feasible} when $A_t^-$ satisfies the critical policy, every committed tail upper bound is known and within SLO, and the stage-wise allocation gap is at most $\varepsilon$.

This certificate is conditional on the uncertainty set containing the relevant performance table. We therefore use “empirical” or “conditionally calibrated,” not an unconditional correctness guarantee.

\subsection{Acquisition}

For each unmeasured experiment $e$, \method estimates how revealing its outcome would shrink capacity/tail uncertainty and change the conservative--optimistic allocation gap:
\[
  \mathrm{score}_t(e)
  = \widehat{\mathbb{E}}\!\left[
       \Gamma_t-\Gamma_{t+1}\mid e
    \right],
\]
where $\Gamma_t$ is the stage-wise downstream allocation gap. The implementation uses a deterministic approximation to this value-of-information objective. Expected profiling GPU-seconds are a strict tie-breaker, not a divisor; consequently we describe the rule as decision-critical rather than universally compute-resource-optimal.

\begin{algorithm}[t]
\caption{\method profiling loop}
\label{alg:fleetsieve}
\begin{algorithmic}[1]
\STATE Initialize a common sparse design for every method.
\WHILE{unmeasured candidates remain}
  \STATE Fit joint capacity and tail intervals.
  \STATE $A^- \gets$ conservative fleet allocation.
  \STATE $A^+ \gets$ optimistic fleet allocation.
  \STATE $z \gets \textsc{Certificate}(A^-,A^+,\varepsilon)$.
  \IF{$z=\textsc{Certified-Feasible}$}
    \RETURN $A^-$.
  \ELSIF{$z=\textsc{Certified-Infeasible}$}
    \RETURN \textsc{Infeasible}.
  \ENDIF
  \STATE Reveal the candidate with maximum decision-gap reduction.
\ENDWHILE
\RETURN the best measured feasible allocation.
\end{algorithmic}
\end{algorithm}

\section{Experimental Methodology}
\label{sec:setup}

\paragraph{System.}
We evaluate a 31B-parameter open-weight decoder model in FP8 on a single H100 node, using a vLLM-based serving stack with speculative decoding disabled. Candidate TP degrees are $\{2,4,8\}$. Each cell runs for 300 seconds; profiling is measured as duration multiplied by GPUs used, reported in GPU-seconds. 

\paragraph{Workloads and SLOs.}
The primary workload uses the public Azure conversation and code traces~\cite{stojkovic2025dynamollm}. Trace records provide arrival times and input/output token counts, not prompt content. The chat policy requires success $\ge99\%$, TTFT-p99 $\le2$ seconds, and completion-p99 $\le30$ seconds. The code policy requires success $\ge99\%$ and TTFT-p99 $\le5$ seconds. Load labels such as C96 are ordinal indices into a documented arrival-scale ladder, not literal in-flight request counts.

The fixed primary grid contains 21 standardized cells plus four earlier code-TP2 calibration cells. The four cells use a different warm-up split and remain separately labeled; excluding them does not change the code oracle because TP2 is dominated. Boundary experiments repeat eight TP4/TP8 cells over five seeds. BurstGPT adds 18 measured cells: three load scales, two TP degrees, and three seeds.

\paragraph{Baselines and metrics.}
All methods start from the same observations, use the same candidate table, SLOs, allocator, and empirical stop. We compare uniform random, fixed grid order, joint maximum uncertainty, a shared-feature uncertainty surrogate, targeted active learning, value of information, constrained Bayesian optimization, and \method. Primary metrics are GPU-seconds to an $\varepsilon$-correct certified decision ($\varepsilon=0.05$), decision regret versus the fully measured oracle, and regret AUC. We also report the resulting 16-GPU allocation, boundary stability, and tail-certificate controls.

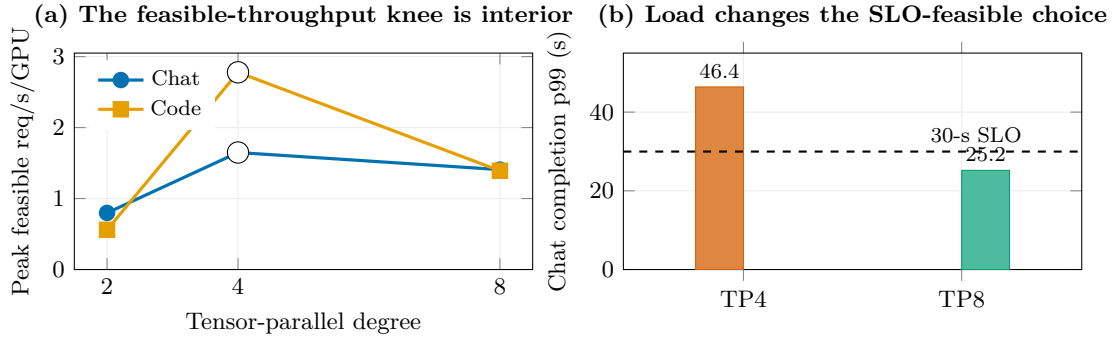
\begin{figure*}[t]
\centering
\begin{tikzpicture}
\begin{groupplot}[
  group style={group size=2 by 1,horizontal sep=1.2cm},
  width=0.43\textwidth,height=0.25\textwidth,
  grid=both,grid style={gray!15},
  tick label style={font=\small},label style={font=\small},
  title style={font=\small\bfseries},legend style={font=\footnotesize,draw=none}
]
\nextgroupplot[
  title={(a) The feasible-throughput knee is interior},
  xlabel={Tensor-parallel degree},ylabel={Peak feasible req/s/GPU},
  xtick={2,4,8},xmin=1.5,xmax=8.5,ymin=0,ymax=3.05,
  legend pos=north west
]
\addplot[color=fsblue,very thick,mark=*,mark size=2.5pt] coordinates {(2,0.798) (4,1.649) (8,1.408)};
\addlegendentry{Chat}
\addplot[color=fsorange,very thick,mark=square*,mark size=2.5pt] coordinates {(2,0.560) (4,2.779) (8,1.392)};
\addlegendentry{Code}
\addplot[only marks,mark=*,mark options={fill=white,draw=black},mark size=4pt] coordinates {(4,1.649) (4,2.779)};

\nextgroupplot[
  title={(b) Load changes the SLO-feasible choice},
  ylabel={Chat completion p99 (s)},
  symbolic x coords={TP4,TP8},xtick={TP4,TP8},
  ymin=0,ymax=55,ybar=0pt,bar width=18pt,
  nodes near coords,nodes near coords style={font=\footnotesize},
  enlarge x limits=0.55
]
\addplot[ybar,fill=fsred!75,draw=fsred] coordinates {(TP4,46.4)};
\addplot[ybar,fill=fsgreen!70,draw=fsgreen] coordinates {(TP8,25.2)};
\draw[black,dashed,thick] (rel axis cs:0,0.545) -- node[pos=0.77,above,font=\footnotesize]{30-s SLO} (rel axis cs:1,0.545);
\end{groupplot}
\end{tikzpicture}
\caption{Why profiling is necessary. (a) Peak SLO-feasible throughput per GPU is non-monotone and peaks at TP4 for both Azure workload classes. (b) At the heavier Chat C128 point, TP4 and TP8 serve the same raw request rate, but only TP8 satisfies the completion-time objective; bars show the five-seed mean.}
\label{fig:knee-tail}
\end{figure*}

\section{Results}
\label{sec:results}

\subsection{The TP optimum is interior and load dependent}

Table~\ref{tab:frontier} and Figure~\ref{fig:knee-tail} report peak SLO-feasible throughput per GPU. Both workload classes peak at TP4; neither smaller nor larger TP is uniformly best.

\begin{table}[t]
\centering
\caption{Peak SLO-feasible throughput (requests/s/GPU).}
\label{tab:frontier}
\begin{tabular}{@{}lccc@{}}
\toprule
Workload & TP2 & TP4 & TP8\\
\midrule
Chat & 0.798 & \textbf{1.649} & 1.408\\
Code & 0.560 & \textbf{2.779} & 1.392\\
\bottomrule
\end{tabular}
\end{table}

Across five repeats at the heavier Chat C128 point, TP4 and TP8 both serve 11.27 requests/s. Their mean completion-p99 values are 46.4 and 25.2 seconds, respectively, so only TP8 passes the 30-second objective. Thus the SLO-feasible TP can change even when throughput does not.

\subsection{Decision-critical profiling is modestly better in aggregate}
\label{sec:profiling}

\begin{figure*}[t]
\centering
\begin{tikzpicture}
\begin{groupplot}[
  group style={group size=2 by 1,horizontal sep=1.25cm},
  width=0.44\textwidth,height=0.27\textwidth,
  symbolic x coords={FS,Rand,Shared,MaxU,Grid,T/V/BO,Full},
  xtick=data,xticklabel style={rotate=28,anchor=east,font=\footnotesize},
  ylabel={GPU-hours to certificate},
  ymin=0,ymax=9,
  grid=major,grid style={gray!15},
  tick label style={font=\small},label style={font=\small},
  title style={font=\small\bfseries},legend style={font=\footnotesize,draw=none}
]
\nextgroupplot[
  title={(a) Aggregate fleet decision},ybar=0pt,bar width=10pt,
  xtick={FS,Rand,Shared,MaxU,Grid,T/V/BO,Full}
]
\addplot[fill=fsblue!80,draw=fsblue] coordinates {(FS,6.167)};
\addplot[fill=fsorange!75,draw=fsorange] coordinates {(Rand,6.623)};
\addplot[fill=fsgray!45,draw=fsgray] coordinates {(Shared,6.833) (MaxU,7.167) (Grid,7.333) (T/V/BO,7.833) (Full,8.167)};
\node[font=\footnotesize,anchor=north west,fill=white,inner sep=2pt] at (rel axis cs:0.02,0.98) {FS: 6.9\% below random};

\nextgroupplot[
  title={(b) The workload split matters},
  symbolic x coords={FS,Rand,Shared,MaxU,Grid,T/V/BO},
  ybar,bar width=7pt,
  legend pos=north west
]
\addplot[fill=fsblue!80,draw=fsblue] coordinates {(FS,2.667) (Rand,3.397) (Shared,4.000) (MaxU,4.000) (Grid,3.833) (T/V/BO,4.333)};
\addlegendentry{Chat}
\addplot[fill=fsorange!75,draw=fsorange] coordinates {(FS,3.500) (Rand,3.227) (Shared,2.833) (MaxU,3.167) (Grid,3.500) (T/V/BO,3.500)};
\addlegendentry{Code}
\end{groupplot}
\end{tikzpicture}
\caption{Profiling GPU-seconds under the common empirical certificate. FS denotes FleetSieve; Rand is uniform random; Shared is the shared-feature uncertainty surrogate; MaxU is joint maximum uncertainty; T/V/BO groups three methods that are equivalent on this grid. FleetSieve is modestly better in aggregate, substantially better for Chat, and not better for Code.}
\label{fig:profiling-resources}
\end{figure*}
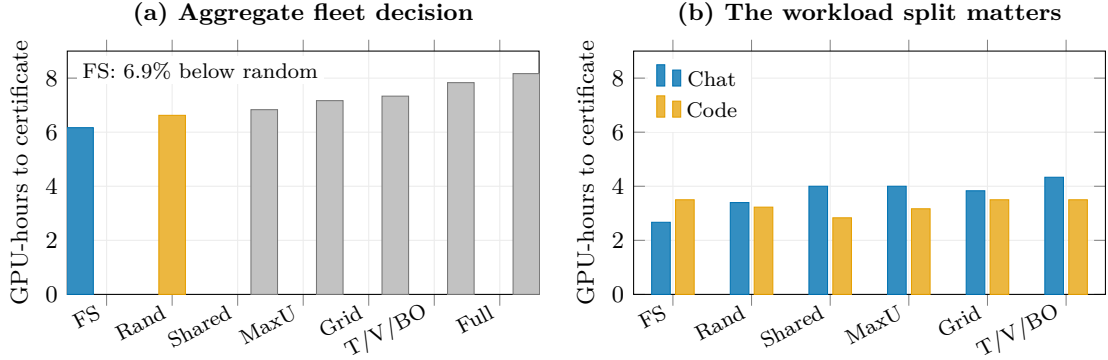

Table~\ref{tab:profiling-resources} and Figure~\ref{fig:profiling-resources} report GPU-seconds to the common empirical certificate. On the fixed aggregate comparison, \method uses 22,200 GPU-seconds, 6.9\% below random and 9.8--21.3\% below the other listed baselines. Every deterministic run and every random trial reaches zero regret; the oracle TP is 4 for both classes.

\begin{table*}[t]
\centering
\caption{GPU-seconds to the same zero-regret certified decision. Lower is better. One row groups three methods that are equivalent on this fixed grid: targeted active learning, value of information, and constrained Bayesian optimization.}
\label{tab:profiling-resources}
\begin{tabular}{@{}lrrr@{}}
\toprule
Method & Aggregate & Chat & Code\\
\midrule
\textbf{\method} & \textbf{22,200} & \textbf{9,600} & 12,600\\
Uniform random & 23,844 & 12,228 & 11,616\\
Shared-feature uncertainty & 24,600 & 14,400 & \textbf{10,200}\\
Joint maximum uncertainty & 25,800 & 14,400 & 11,400\\
Fixed grid order & 26,400 & 13,800 & 12,600\\
Targeted AL / value-VOI / constrained BO & 28,200 & 15,600 & 12,600\\
Full grid & 29,400 & -- & --\\
\bottomrule
\end{tabular}
\end{table*}

The class split is important. Against random, the fixed Chat comparison saves 21.5\%; against the remaining baselines it saves 30.4--38.5\%. Its Chat regret AUC is 0.1256, about half the fixed-grid value of 0.2459. For Code, random, shared-feature uncertainty, and joint uncertainty use fewer GPU-seconds than \method because TP4 becomes evident early.

The table value and the randomized estimate answer different questions. Across 200 random reveal orders, the ratio-of-means advantage over random is 5.4\%, with a 95\% bootstrap CI of 3.5--7.2\%. For Chat it is 19.4\% (16.2--22.3\%). On a single random draw, \method uses fewer GPU-seconds only 58\% of the time in aggregate and 65\% for Chat. The gain is reliable in expectation, not for every ordering. A 500-replicate parametric bootstrap over the repeated boundary cells never changes the oracle decision; this noise analysis covers the repeated cells and holds the remaining grid cells fixed.

A heuristic early stop based on a fixed $1.5\times$ ceiling is unsafe: fixed-grid and joint-uncertainty profiling stop with TP8 on Chat and regret 0.2407. Under the common empirical certificate, all methods continue to the correct TP4 frontier. This failure motivates evaluating the stopping rule rather than hindsight “first correct” GPU-seconds.

\subsection{Profiling changes a constrained 16-GPU allocation}

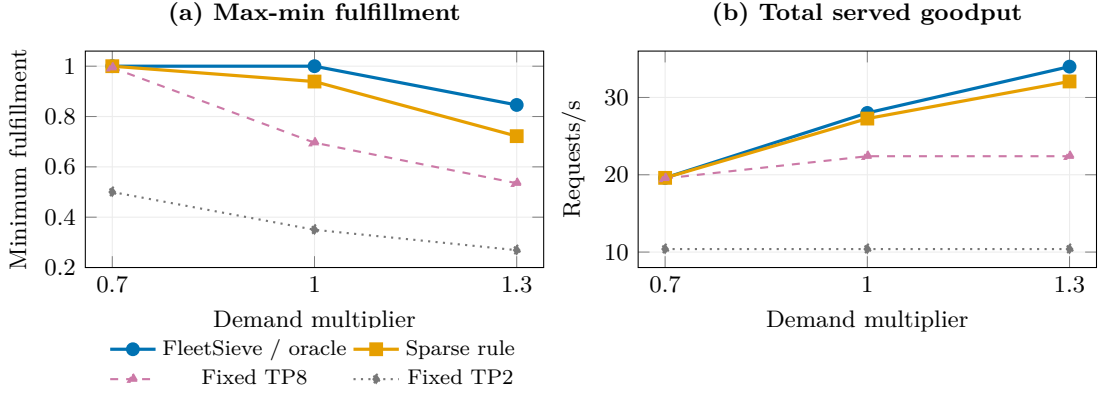
\begin{figure*}[t]
\centering
\begin{tikzpicture}
\begin{groupplot}[
  group style={group size=2 by 1,horizontal sep=1.25cm},
  width=0.43\textwidth,height=0.25\textwidth,
  xlabel={Demand multiplier},xtick={0.7,1.0,1.3},xmin=0.66,xmax=1.34,
  grid=both,grid style={gray!15},
  tick label style={font=\small},label style={font=\small},
  title style={font=\small\bfseries},
  legend style={font=\footnotesize,draw=none,at={(0.5,-0.28)},anchor=north,legend columns=2}
]
\nextgroupplot[title={(a) Max-min fulfillment},ylabel={Minimum fulfillment},ymin=0.2,ymax=1.06]
\addplot[color=fsblue,very thick,mark=*] coordinates {(0.7,1.000) (1.0,1.000) (1.3,0.846)};
\addlegendentry{FleetSieve / oracle}
\addplot[color=fsorange,very thick,mark=square*] coordinates {(0.7,1.000) (1.0,0.939) (1.3,0.722)};
\addlegendentry{Sparse rule}
\addplot[color=fspurple,thick,dashed,mark=triangle*] coordinates {(0.7,0.994) (1.0,0.696) (1.3,0.535)};
\addlegendentry{Fixed TP8}
\addplot[color=fsgray,thick,dotted,mark=diamond*] coordinates {(0.7,0.500) (1.0,0.350) (1.3,0.269)};
\addlegendentry{Fixed TP2}

\nextgroupplot[title={(b) Total served goodput},ylabel={Requests/s},ymin=8,ymax=36]
\addplot[color=fsblue,very thick,mark=*] coordinates {(0.7,19.60) (1.0,28.00) (1.3,33.99)};
\addplot[color=fsorange,very thick,mark=square*] coordinates {(0.7,19.60) (1.0,27.27) (1.3,32.07)};
\addplot[color=fspurple,thick,dashed,mark=triangle*] coordinates {(0.7,19.53) (1.0,22.40) (1.3,22.40)};
\addplot[color=fsgray,thick,dotted,mark=diamond*] coordinates {(0.7,10.39) (1.0,10.39) (1.3,10.39)};
\end{groupplot}
\end{tikzpicture}
\caption{A profiling decision changes the resource-coupled fleet outcome. At low demand, spare capacity hides the difference. As demand rises, the sparse TP8/TP4 choice loses both fairness and served load relative to FleetSieve and the fully measured oracle. Every method uses the same 16-GPU resource allocation.}
\label{fig:fleet-allocation}
\end{figure*}

We allocate 16 GPUs between Chat and Code at three demand scales (Table~\ref{tab:allocation} and Figure~\ref{fig:fleet-allocation}). The fully measured oracle and \method choose two TP4 replicas per class. A sparse independent rule chooses one TP8 Chat replica and two TP4 Code replicas. Both consume eight GPUs per class, but the Chat capacity differs.

\begin{table*}[t]
\centering
\caption{Effect of the profiling decision on the 16-GPU allocation. MM is max-min fulfillment; goodput is served requests/s. The sparse rule chooses Chat TP8 rather than the oracle TP4.}
\label{tab:allocation}
\begin{tabular}{@{}clccrr@{}}
\toprule
Demand & Method & TP (Chat/Code) & Replicas & MM & Goodput\\
\midrule
0.7$\times$ & Oracle / \method & 4 / 4 & 2 / 2 & 1.000 & 19.60\\
             & Sparse rule & 8 / 4 & 1 / 2 & 1.000 & 19.60\\
\midrule
1.0$\times$ & Oracle / \method & 4 / 4 & 2 / 2 & 1.000 & 28.00\\
             & Sparse rule & 8 / 4 & 1 / 2 & 0.939 & 27.27\\
             & Fixed TP8 & 8 / 8 & 1 / 1 & 0.696 & 22.40\\
             & Fixed TP2 & 2 / 2 & 3 / 5 & 0.350 & 10.39\\
\midrule
1.3$\times$ & Oracle / \method & 4 / 4 & 2 / 2 & 0.846 & 33.99\\
             & Sparse rule & 8 / 4 & 1 / 2 & 0.722 & 32.07\\
\bottomrule
\end{tabular}
\end{table*}

At 0.7$\times$ demand, spare capacity masks the decision and all three serve the full load. At 1.0$\times$, the sparse decision loses 0.73 requests/s and 6.1 percentage points of max-min fulfillment. At 1.3$\times$, the losses grow to 1.93 requests/s and 12.4 points. This is the resource-coupled consequence that an isolated TP-accuracy metric misses. Because every row uses the same 16 GPUs, we do not report a GPU-count saving.

\subsection{Tail latency is load-bearing in the certificate}
\label{sec:certificate}

A capacity-only rule selects Chat TP4 at C128 because it serves 11.27 requests/s, but its 46.4-second completion-p99 violates the 30-second objective. The joint rule selects TP4 at C96 for the frontier decision, where completion-p99 is 15.98 seconds, and uses TP8 when C128 must be served. The certificate refuses to fire until the committed configuration's tail upper bound is known.

A separate fixed-order audit exercises the state machine. A non-critical trajectory certifies at step 20 (27,000 GPU-seconds); a trajectory with critical floors of 6 requests/s for Chat and 10 requests/s for Code certifies at step 17 (22,800 GPU-seconds); an intentionally infeasible 1,000-requests/s demand returns \textsc{Certified-Infeasible}. These GPU-seconds audit the certificate and are not the acquisition-comparison GPU-seconds in Table~\ref{tab:profiling-resources}.

The initial offline audit used an optimistic prior derived from unobserved peak cells early in the trajectory. To test whether it affected the result, we replaced that branch with a revealed-only prior and replayed the audit. The branch is inactive after the initial six probes, and both versions certify at the same step, GPU-seconds, and allocation. We label the certificate empirical and conditionally calibrated; an online implementation should use the revealed-only path.

\subsection{Repeated boundaries and a second public trace}
\label{sec:breadth}

Across five seeds on eight boundary cells, seven cells retain the same feasibility outcome. Chat TP4 at C96 is the exception: it satisfies the policy in four of five runs because one run falls to 94.3\% success. The load-bearing C128 distinction is stable in every run: TP4 fails with a mean completion-p99 of 46.38 seconds, while TP8 passes at 25.15 seconds.

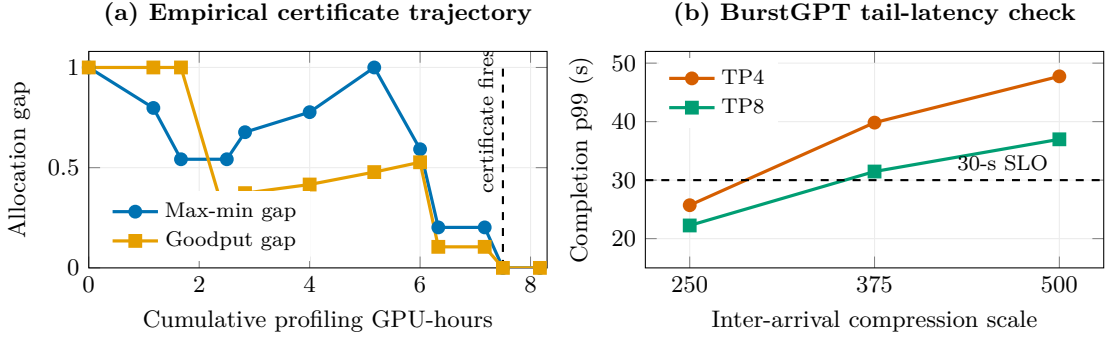
\begin{figure*}[t]
\centering
\begin{tikzpicture}
\begin{groupplot}[
  group style={group size=2 by 1,horizontal sep=1.3cm},
  width=0.43\textwidth,height=0.25\textwidth,
  grid=both,grid style={gray!15},
  tick label style={font=\small},label style={font=\small},
  title style={font=\small\bfseries},legend style={font=\footnotesize,draw=none}
]
\nextgroupplot[
  title={(a) Empirical certificate trajectory},
  xlabel={Cumulative profiling GPU-hours},ylabel={Allocation gap},
  xmin=0,xmax=8.3,ymin=0,ymax=1.08,legend pos=south west
]
\addplot[color=fsblue,very thick,mark=*] coordinates {
 (0,1.000) (1.17,0.798) (1.67,0.542) (2.50,0.542)
 (2.83,0.677) (4.00,0.777) (5.17,1.000) (6.00,0.592)
 (6.33,0.202) (7.17,0.202) (7.50,0.000) (8.17,0.000)};
\addlegendentry{Max-min gap}
\addplot[color=fsorange,very thick,mark=square*] coordinates {
 (0,1.000) (1.17,1.000) (1.67,1.000) (2.50,0.228)
 (2.83,0.373) (4.00,0.416) (5.17,0.478) (6.00,0.527)
 (6.33,0.105) (7.17,0.105) (7.50,0.000) (8.17,0.000)};
\addlegendentry{Goodput gap}
\draw[black,dashed,thick] (axis cs:7.5,0) -- node[pos=0.72,rotate=90,above,font=\footnotesize]{certificate fires} (axis cs:7.5,1.02);

\nextgroupplot[
  title={(b) BurstGPT tail-latency check},
  xlabel={Inter-arrival compression scale},ylabel={Completion p99 (s)},
  xtick={250,375,500},xmin=220,xmax=530,ymin=15,ymax=52,legend pos=north west
]
\addplot[color=fsred,very thick,mark=*] coordinates {(250,25.712) (375,39.831) (500,47.742)};
\addlegendentry{TP4}
\addplot[color=fsgreen,very thick,mark=square*] coordinates {(250,22.251) (375,31.472) (500,36.979)};
\addlegendentry{TP8}
\draw[black,dashed,thick] (axis cs:220,30) -- node[pos=0.78,above,font=\footnotesize]{30-s SLO} (axis cs:530,30);
\end{groupplot}
\end{tikzpicture}
\caption{Uncertainty and robustness. (a) A fixed-order audit shows that the conservative and optimistic fleet objectives converge only after the decisive tail measurement; the non-monotonic gaps reflect changes in the optimistic allocation as new cells are revealed. (b) On BurstGPT, TP8 has lower completion p99 at every tested load, but the three-point grid does not contain a TP8-only feasible point.}
\label{fig:certificate-burst}
\end{figure*}

BurstGPT reproduces the mechanism direction: TP8 has lower completion-p99 at all three tested loads, and the gap grows with load (Figure~\ref{fig:certificate-burst}). It does \emph{not} reproduce a uniquely feasible TP8 cell: both TPs pass at scale 250 and both fail at 375. The trace's absolute timescale is compressed 250--500$\times$, so this is a load-controlled robustness check, not a timescale-faithful operational replay or a second end-to-end acquisition win.

\section{Discussion and Limitations}
\label{sec:limitations}

The evaluation covers one 31B-parameter model, one serving stack, and one H100 platform. Model architecture, quantization, interconnect, and runtime can change the balance between collective-communication time and KV-cache capacity, and may move or remove the observed interior optimum. BurstGPT broadens the token-length and burst distribution, but tests the tail-latency mechanism rather than the complete acquisition algorithm.

The certificate relies on empirically calibrated residual bands and an interior-knee structural assumption. A multimodal or sharply discontinuous capacity curve could violate these assumptions. Measurement noise also matters near an SLO boundary: one of eight repeated cells changes feasibility across seeds, although the C128 cell that changes the TP choice is stable in every repeat.

The primary grid includes four TP2 calibration cells collected with a different warm-up split. They remain labeled separately, and removing them does not change the oracle because TP2 is dominated on the reported metric. Finally, the fleet experiment fixes the GPU resource allocation at 16. It improves served demand and fairness without reducing the steady-state GPU count.

\section{Conclusion}

\method profiles an LLM serving configuration space according to the fleet allocation that the measurements inform. On the evaluated H100 grid, it reaches the oracle aggregate decision using 22,200 GPU-seconds, 6.9\% less than uniform random profiling in the fixed comparison. The benefit is larger for Chat and absent for Code. Joint capacity and tail modeling prevents a measured 46.4-second SLO violation, while the resulting TP choice recovers up to 1.93 requests/s and 12.4 percentage points of max-min fulfillment in a 16-GPU fleet. These results show that decision-aware profiling can reduce measurement effort when uncertainty affects the allocation, while its value remains workload- and calibration-dependent.

\appendix

\section{Reproducibility Protocol}

Each standardized primary cell uses a 30-second warm-up and a 270-second measurement window. The four earlier calibration cells use a 10-second warm-up and 290-second measurement window and are never silently mixed with standardized repeats. Run metadata include model precision, TP, trace window, arrival scale, warm-up, measurement duration, seed, success rate, TTFT, completion time, and failure state. Raw records are immutable; derived feasibility is recomputed from the class policy.

The initial observations, candidate set, per-cell GPU-seconds, allocator, SLO policy, and oracle table are identical across all profiling methods. Random profiling is evaluated over multiple reveal orders. Failed, rejected, and timed-out requests do not count as SLO goodput.

\section{Baseline Definitions}

\begin{itemize}[leftmargin=*]
\item \textbf{Uniform random} samples remaining cells uniformly.
\item \textbf{Fixed grid} traverses a predefined TP/load order.
\item \textbf{Joint maximum uncertainty} selects the largest marginal uncertainty from the shared model without decision impact.
\item \textbf{Shared-feature uncertainty} uses an RBF cross-TP surrogate and selects by uncertainty.
\item \textbf{Targeted active learning} prioritizes information about the current decision.
\item \textbf{Value of information} estimates expected decision improvement with its own compute-resource normalization.
\item \textbf{Constrained Bayesian optimization} uses independent per-arm models, expected improvement, and feasibility constraints.
\end{itemize}


\begin{thebibliography}{99}

\bibitem{kwon2023vllm}
W.~Kwon et al.
Efficient memory management for large language model serving with PagedAttention.
In \emph{Proceedings of SOSP}, 2023.

\bibitem{yu2022orca}
G.-I.~Yu et al.
Orca: A distributed serving system for transformer-based generative models.
In \emph{Proceedings of OSDI}, 2022.

\bibitem{agrawal2024sarathi}
A.~Agrawal et al.
Sarathi-Serve: Efficient LLM serving with chunked prefills.
In \emph{Proceedings of OSDI}, 2024.

\bibitem{zhong2024distserve}
Y.~Zhong et al.
DistServe: Disaggregating prefill and decoding for goodput-optimized large language model serving.
In \emph{Proceedings of OSDI}, 2024.

\bibitem{patel2024splitwise}
P.~Patel et al.
Splitwise: Efficient generative LLM inference using phase splitting.
In \emph{Proceedings of ISCA}, 2024.

\bibitem{li2023alpaserve}
Z.~Li et al.
AlpaServe: Statistical multiplexing with model parallelism for deep learning serving.
In \emph{Proceedings of OSDI}, 2023.

\bibitem{chen2025slosserve}
S.~Chen, Z.~Jia, S.~Khan, A.~Krishnamurthy, and P.~B. Gibbons.
SLOs-Serve: Optimized serving of multi-SLO LLMs.
arXiv:2504.08784, 2025.

\bibitem{srivatsa2026nitsum}
V.~Srivatsa, Z.~He, P.~Guo, D.~Li, and Y.~Zhang.
Nitsum: Serving tiered LLM requests with adaptive tensor parallelism.
arXiv:2605.05467, 2026.

\bibitem{cheng2025scoot}
K.~Cheng, Z.~Wang, W.~Hu, T.~Yang, J.~Li, and S.~Zhang.
SCOOT: SLO-oriented performance tuning for LLM inference engines.
In \emph{Proceedings of The Web Conference}, 2025.

\bibitem{settles2012active}
B.~Settles.
\emph{Active Learning}.
Morgan \& Claypool, 2012.

\bibitem{filstroff2021targeted}
L.~Filstroff, I.~Sundin, P.~Mikkola, A.~Tiulpin, J.~Kylm{\"a}oja, and S.~Kaski.
Targeted active learning for Bayesian decision-making.
arXiv:2106.04193, 2021.

\bibitem{nakamura2023combgape}
S.~Nakamura and M.~Sugiyama.
A fast algorithm for the real-valued combinatorial pure exploration of multi-armed bandit.
arXiv:2306.09202, 2023.

\bibitem{stojkovic2025dynamollm}
J.~Stojkovic, C.~Zhang, I.~Goiri, J.~Torrellas, and E.~Choukse.
DynamoLLM: Designing LLM inference clusters for performance and energy efficiency.
In \emph{Proceedings of HPCA}, 2025.

\bibitem{wang2024burstgpt}
Y.~Wang et al.
BurstGPT: A real-world workload dataset to optimize LLM serving systems.
In \emph{Proceedings of KDD}, 2025.

\bibitem{cheng2014kcst}
H.~Cheng, X.~Fei, M.~Almulla, and A.~Boukerche.
A knapsack constrained Steiner tree model for continuous coverage over urban VANETs.
In \emph{Proceedings of IEEE ICC}, pages 130--135, 2014.

\end{thebibliography}
\end{document}